\documentclass[runningheads]{llncs}

\usepackage{eccv}

\usepackage{eccvabbrv}

\usepackage{graphicx}
\usepackage{booktabs}
\usepackage{multirow}
\usepackage{adjustbox}
\usepackage{pifont}

\usepackage[accsupp]{axessibility}  

\usepackage[pagebackref,breaklinks,colorlinks,citecolor=eccvblue]{hyperref}

\usepackage{orcidlink}

\begin{document}

\title{TextDS: Parameter-Efficient Representation Alignment for Scene Text Detection under Distribution Shifts} 

\titlerunning{TextDS: Scene Text Detection under Distribution Shifts}

\author{
Boyuan Chen\inst{1}\textsuperscript{*}\and
Zichen Dang\inst{2}\textsuperscript{*}\and
Chuang Yang\inst{2}\and \\
Lap-Pui Chau\inst{2}\and
Yi Wang\inst{2}\textsuperscript{$\dagger$}
}

\authorrunning{B. Chen et al.}

\institute{School of Electrical Engineering, Xi'an Jiaotong University \and
Department of Electrical and Electronic Engineering, \\The Hong Kong Polytechnic University \\
\email{15956368@stu.xjtu.edu.cn}, \email{zichen.dang@connect.polyu.hk}
\email{\{c1yang,lap-pui.chau,yi-eie.wang\}@polyu.edu.hk}}

\maketitle
\renewcommand{\thefootnote}{} 
\footnotetext{
    \textsuperscript{*} Equal Contribution \quad
    \textsuperscript{\textdagger} Corresponding Author.
}

\setcounter{footnote}{0}
\renewcommand{\thefootnote}{\arabic{footnote}} 

\begin{abstract}
  In real-world deployments, scene text detectors inevitably face distribution shifts beyond the training distribution. Prior work often depends on large-scale scene-text pretraining, yet evaluation under cross-domain changes and real-world imaging degradations remains limited. We propose TextDS, an efficient framework for scene text detection under distribution shifts. First, we propose a data-efficient dual-encoder design with visual foundation models, eliminating the reliance on large-scale scene-text pretraining. Second, we introduce Step-wise LoRA adaptation (SWLoRA), which performs progressive low-rank refinement with a dynamic early-exit mechanism for effective feature adaptation. Third, we propose Common Subspace Fusion (CSF) to align and fuse the two branches in a shared subspace while retaining complementary, shift-robust information. Finally, we construct adverse-condition scene text detection datasets to address the gap in evaluating under imaging degradation. Experiments show that TextDS achieves competitive performance in scene text detection, demonstrating robustness across domains and adverse imaging conditions with only 4.9M trainable parameters. The code is publicly available at \url{https://github.com/ZChenDang/TextDS}
\end{abstract}

\section{Introduction}

Scene text detection aims to localize text instances in natural scenes, and serves as a foundation for end-to-end OCR and downstream understanding \cite{naiemi2022scene}. Reliable scene text detection enables a wide range of real-world applications, including instant translation and navigation \cite{vaidya2024show}, mobile document capture and payment \cite{neat2019scene}, industrial inspection \cite{guan2022industrial}, and embodied robotic perception \cite{song2024scene}. As the scene text often carries dense, explicit semantic cues, the quality of detection directly sets the ceiling for subsequent recognition and parsing \cite{long2021scene, wang2021self, wang2021rethinking}. With increasing demand for large-scale, low-latency, and cross-device deployment, developing methods that are not only accurate but also stable and deployment-friendly is of significant scientific and practical importance \cite{luo2024delving, wang2020convolutional}.

\begin{figure}[tb]
  \centering
  \includegraphics[height=4.2cm]{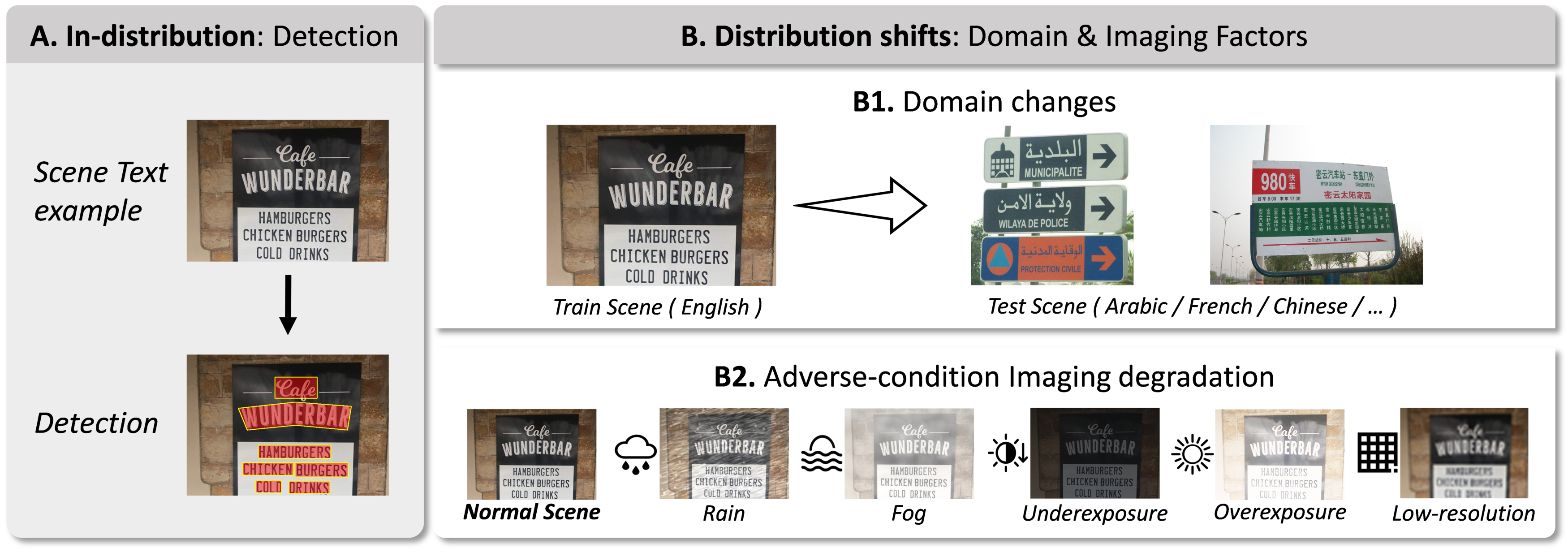}
  \caption{Examples of distribution shifts in scene text detection, including domain changes and adverse-condition imaging degradation. The example of domain changes is transformation of text language from English to French, Arabic and Chinese. Imaging degradation includes low-resolution, rain, fog, underexposure and overexposure.
  }
  \label{fig1}
\end{figure}

Current scene text detection methods can be categorized into two groups, including pixel aggregation approaches \cite{liao2020real, zhang2022arbitrary, wang2025s3inet} and explicit structure modeling approaches \cite{zhang2021adaptive, su2024lranet, su2025explicit}. Overall, pixel aggregation and explicit structure modeling represent two complementary paradigms for scene text detection, emphasizing dense prediction with instance aggregation and direct instance-level structure prediction, respectively.

However, existing methods are developed under relatively ideal acquisition conditions \cite{yuliang2017detecting, ch2017total}, and rely heavily on large-scale pre-training on datasets such as SynthText-800k or SynthText-150k \cite{gupta2016synthetic} to improve in-distribution performance, leaving evaluation under distribution shifts insufficiently studied and rarely optimized. In real-world scene deployments, scene text detectors encounter distribution shifts relative to the training distribution \cite{luo2024grounding}, driven by two types: \textbf{cross-domain changes in data sources} and \textbf{imaging degradations introduced during capture and transmission} \cite{yang2024pragmatic}, which is illustrated in \cref{fig1}. The former type of domain shift involves changes in distribution caused by differences in data sources, such as variations in shooting scenes, font languages, and font styles, which result in the actual application data distribution deviating from the training distribution \cite{xue2022language}. For the latter type, real imaging degradations are pervasive and physically grounded \cite{jiang2025survey}. For instance, rain and fog reduce contrast through scattering, low-light and underexposure compress dynamic range and increase noise \cite{brophy2023review, arif2022comprehensive}, strong illumination or exposure failures cause overexposure and highlight saturation that erase stroke and boundary details \cite{zhao2025deep}, and long-range capture, digital zoom, and compressed transmission produce low-resolution inputs with lost fine textures\cite{su2025review}. Such shifts simultaneously undermine pixel separability and geometric cues, thereby destabilizing thresholding and region growing in pixel-aggregation methods and weakening boundary localization and component association in structure-modeling methods. Despite this, specific datasets and systematic evaluations for these adverse conditions remain scarce, hindering robust assessment under unified distribution-shift settings. 

To address the aforementioned issues, we propose TextDS, an efficient scene text detection network designed for distribution-shift settings. TextDS is to achieve both data-efficient and parameter-efficient detection with only a small number of trainable parameters, without relying on large-scale scene-text-specific pretraining, while improving cross-domain generalization and enabling robust evaluation under degraded imaging conditions. The main contributions of this paper are summarized as follows:
\begin{itemize}
    \item \textbf{Dual-branch feature extraction and fusion.} We propose a dual-branch architecture on complementary visual foundation models for scene text feature extraction, and propose Common Subspace Fusion (CSF) to align and fuse the two branches in a shared subspace, avoiding reliance on large-scale scene-text-specific pretraining, with a small trainable parameter count.

    \item \textbf{Dynamic step-wise low-rank adaptation.} We propose Step-wise Low-rank Adaptation (SWLoRA) with a cosine-similarity based dynamic early-exit mechanism, enabling progressive refinement and adaptive stopping of the step-wise updates across samples.
    
    \item \textbf{Extensive evaluation across domains and degradations.} We validate our method with extensive experiments across multiple domains, and complement standard evaluation with proposed adverse-condition datasets to facilitate robustness analysis under diverse degradations.
\end{itemize}

\section{Related Work}

\subsection{Scene Text Detection Methods}

We briefly review representative scene text detection approaches from two complementary perspectives: pixel aggregation and explicit structure modeling. 

Pixel aggregation approaches are centered on dense prediction, such as probability maps, kernel or center regions, and direction or distance fields, and then aggregate pixels into text instances via binarization, connected components, and region growing. Liao \etal~\cite{liao2020real} proposed DB-Net, which makes binarization differentiable and learns adaptive thresholds to reduce reliance on handcrafted thresholds and complicated aggregation rules. Building on the idea of improving supervision and reconstruction robustness, Zhang \etal~\cite{zhang2022arbitrary} proposed TextPMs to address uncertainty introduced by coarse-grained annotations by using a set of probability maps to model text-pixel probability distributions and then reconstructing instances via iterative prediction and region growing. Following this direction, Wang \etal~\cite{wang2025s3inet} proposed S3INet, which strengthens multi-scale semantics and foreground prominence via semantic spatial interaction on single-level features, and suppress text-like backgrounds, thereby reducing false positives. 

In contrast, explicit structure modeling approaches treat instance structure as the direct prediction target, and incorporate boundary constraints or component relations into end-to-end learning to reduce dependence on heuristic aggregation. Along this line, Zhang \etal~\cite{zhang2021adaptive} proposed TextBPN, which refines contour details via boundary proposals and iterative deformation. From a compact shape representation perspective, Su \etal~\cite{su2024lranet} proposed LRANet, which learns data-driven shape bases via low-rank approximation and reconstructs text contours using a small number of coefficients to balance accuracy and efficiency. Moving from contour parameterization to component relation modeling, Su \etal~\cite{su2025explicit} proposed ERRNet, which decomposes text into an ordered sequence of components and explicitly models inter-component relations from a tracking perspective, enabling component-level inference without post-processing. Overall, while pixel aggregation methods achieve instance formation through dense predictions and subsequent grouping, explicit structure modeling methods aim to predict instance geometry or component relations directly.

In addition, in recent years, end-to-end OCR based on large language models has also been introduced into the scene text detection and recognition process, gradually shifting from the detection paradigm to stronger general visual understanding capabilities. Qwen-OCR \cite{bai2025qwen3}, Deepseek-OCR \cite{wei2025deepseek}, and PP-OCR \cite{li2022pp} provide more engineering-oriented and readily deployable solutions. Overall, the large language model-based approach has enhanced the semantic understanding ability for complex scene texts, while the scene text detection methods still have competitiveness in terms of real-time performance and controllability. Both are also showing a trend of integration.

\subsection{Foundation Models: Representation, Segmentation, and Efficient Adaptation}

Vision foundation models offer strong general-purpose priors that can be transferred to downstream tasks with limited task-specific data. In particular, DINO \cite{caron2021emerging} learns Vision Transformer representations through self-distillation without manual annotations, producing features with strong semantic aggregation that transfer well to diverse tasks \cite{li2023mask, ayzenberg2024DINOv2, jose2025DINOv2}. Building on this line, DINOv2 \cite{oquab2023DINOv2} and DINOv3 \cite{simeoni2025dinov3} further advance the pretraining strategy and scaling to yield even stronger transferability. For scene text detection, prior practice often relies on large-scale text-specific synthetic pretraining such as SynthText \cite{gupta2016synthetic}. In contrast, DINOv3 provides an alternative path where general self-supervised weights can be adapted to text detection with lightweight task-specific components.

Alongside strong representations, promptable segmentation models further broaden the applicability of foundation models. SAM \cite{kirillov2023segment} and SAM2 \cite{ravi2024sam} decouple image encoding from mask prediction and generate class-agnostic masks conditioned on prompts (e.g., points and boxes), enabling reuse of dense representations across tasks. SAM-style models have been adopted in domains beyond natural images, including medical imaging \cite{chen2024ma, miao2026sam}, remote sensing \cite{yan2023ringmo, ma2024sam}, and video tracking \cite{xiong2025efficient}. However, their systematic exploration in scene text detection remains limited.

To make adapting large foundation models feasible under limited compute and data, LoRA \cite{hu2022lora} injects low-rank trainable updates into pretrained linear layers while freezing the backbone, enabling effective transfer with a small number of additional parameters and reduced training cost \cite{mao2025survey}. Recent variants such as AdaLoRA \cite{zhang2023adalora} and DoRA \cite{liu2024dora} further improve adaptation quality while keeping the updates lightweight, yet LoRA-style designs tailored to the specific challenges of scene text detection remain underexplored.

\begin{figure}[tb]
  \centering
  \includegraphics[height=8.8cm]{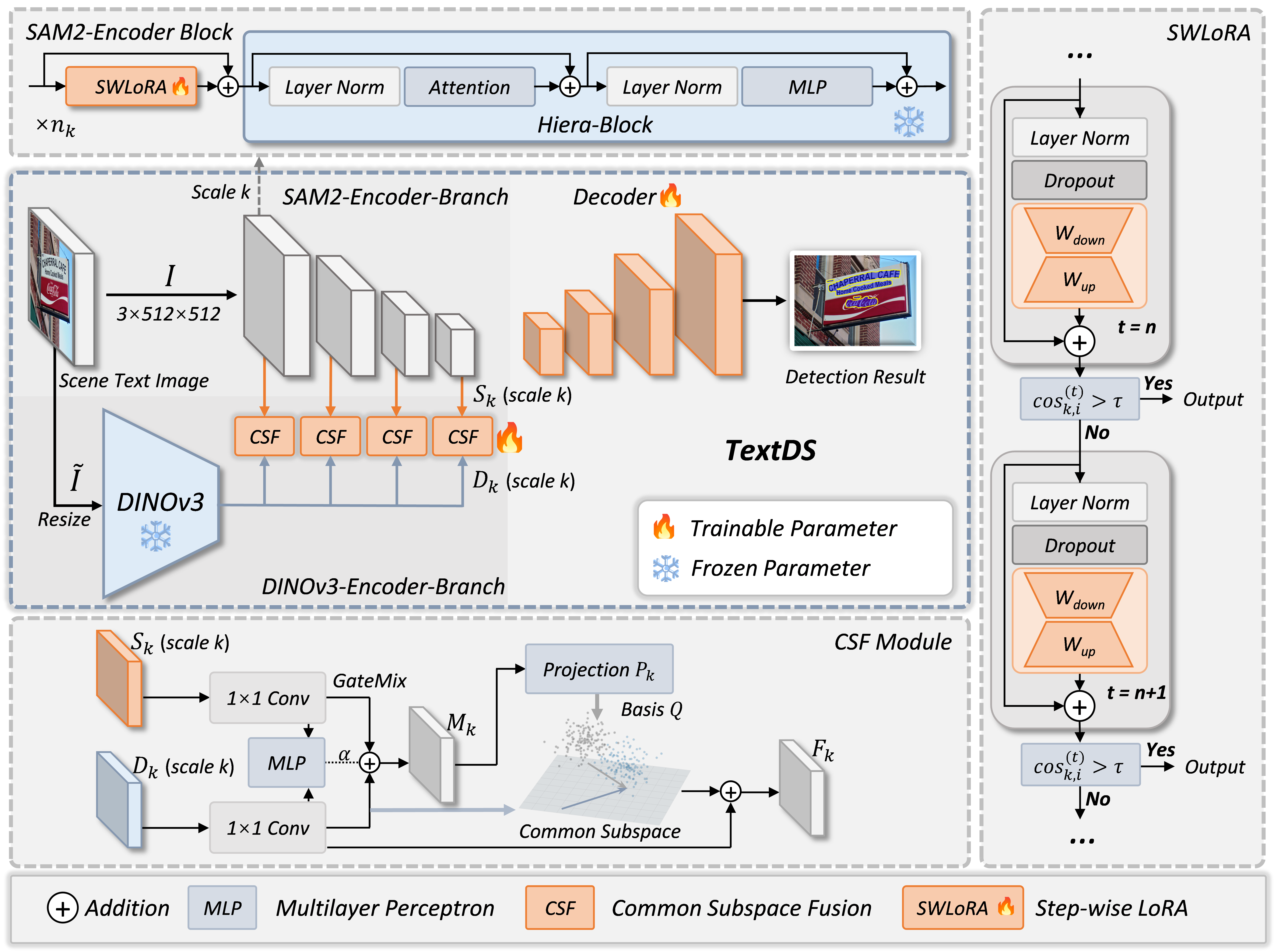}
  \caption{Overall structure of TextDS. The input scene text image is processed through the SAM2-Encoder-Branch and the DINOv3-Encoder-Branch, and the dual-branch encoded features of multiple scales are fused through the Common Subspace Fusion (CSF) module, and each SAM2 Encoder Block uses Step-wise LoRA (SWLoRA) structure for fine-tuning.
  }
  \label{fig2}
\end{figure}

\section{Method}

\subsection{Overview of TextDS Architecture}

Our proposed TextDS follows an overall approach of constructing a dual-encoder using data-efficient representations of the source (SAM2 and DINOv3) that do not rely on scene text for large-scale pre-training. Then, through a parameter-efficient adaptation and fusion mechanism, the cross-domain robust general representation is transformed into the discriminative features required for scene text detection. Specifically, the SAM2 encoding branch provides strong structure and multi-scale priors, while the DINOv3 encoding branch provides robust semantic representation. We propose Step-wise LoRA (SWLoRA) at the block-level input of the SAM2 backbone to complete task adaptation, and propose Common Subspace Fusion (CSF) to fuse the two branches within the shared subspace, while retaining the domain-robust information of DINOv3 in the orthogonal complement space. Finally, we output the text probability map with a lightweight decoder. The overall structure is shown in \cref{fig2}.

\subsection{SAM2-DINOv3 Dual-Encoder Architecture}
To achieve robust scene text detection under distribution shifts while maintaining a data-efficient representation, we adopt a dual-encoder architecture composed of SAM2-Hiera-L and DINOv3 ViT-L/16. Given an input image
\begin{equation}
I \in \mathbb{R}^{B \times 3 \times H \times W},
\end{equation}
where $B$ denotes the batch size and $H \times W$ is the input resolution, the SAM2 encoding branch produces a four-level feature pyramid capturing strong structural priors, while the DINOv3 branch provides a high-level, domain-robust semantic representation. We then align DINOv3 features to each SAM2 scale for subsequent cross-branch fusion and decoding.

\subsubsection{SAM2-Hiera-L for multi-scale structural encoding.}
We keep only the SAM2 image encoder trunk (removing the prompt encoder, mask decoder, and memory-related components), as shown in \cref{fig2}, and use it as a frozen structural feature extractor. The SAM2 trunk yields four-scale features
\begin{equation}
\{S_1,S_2,S_3,S_4\}=E_{\mathrm{sam2}}(I),\qquad 
S_k\in\mathbb{R}^{B\times C_k\times H_k\times W_k},\ k\in\{1,2,3,4\},
\end{equation}
where $E_{\mathrm{sam2}}(\cdot)$ denotes the SAM2-Hiera-L trunk, $S_k$ is the output at the $k$-th scale, and $C_k$, $H_k$, $W_k$ are its channel dimension and spatial resolution, respectively. In our implementation, the channel dimensions are fixed to
\begin{equation}
(C_1,C_2,C_3,C_4)=(144,\ 288,\ 576,\ 1152).
\end{equation}

These features form a pyramid from high to low resolution, following the hierarchical downsampling pattern. For a consistent description of hierarchical computation, we name the four scales as Hiera-S1/S2/S3/S4 and denote the set of Hiera blocks in the $k$-th scale by
\begin{equation}
\mathcal{B}_k=\{b_{k,1},\dots,b_{k,n_k}\},\qquad k\in\{1,2,3,4\},
\label{eq:k_scale}
\end{equation}
where $b_{k,i}$ is the $i$-th block at scale $k$ and $n_k$ is the number of blocks in that scale, and the total number of blocks equals $\sum_{k=1}^4 n_k$. In our model, each trunk block is wrapped by a parameter-efficient adapter (introduced in the next subsection) while the original SAM2 parameters remain frozen, enabling task adaptation with minimal trainable parameters.

\subsubsection{DINOv3 ViT-L/16 semantic extraction and multi-scale alignment.}
For the DINOv3 encoding branch, we resize the input to a fixed resolution $448\times 448$:
\begin{equation}
\tilde I=\mathrm{Resize}(I,448,448).
\end{equation}

We take the last output feature of DINOv3 (the last element returned by a features\_only interface), denoted as
\begin{equation}
D=E_{\mathrm{DINO}}^{(\mathrm{last})}(\tilde I),\qquad 
D\in\mathbb{R}^{B\times 1024\times 28\times 28}.
\end{equation}
Here, $E_{\mathrm{DINO}}^{(\mathrm{last})}$ represents the final semantic representation of ViT-L/16. The spatial size $28\times 28$ comes from the patch size of $16$ (i.e., $448/16=28$), and the channel dimension is $1024$.

Since $D$ is a single-scale feature map, we map and align it to each SAM2 scale via four $1\times1$ convolutions followed by bilinear resizing:
\begin{equation}
D_k=\mathrm{Resize}\!\left(Align_k(D),(H_k,W_k)\right),    Align_k:\mathbb{R}^{1024}\rightarrow\mathbb{R}^{C_k}.
\end{equation}

This produces a scale-aligned pair $(S_k,D_k)$ at each sampling level, where $C_k\in\{144,288,576,1152\}$. The aligned multi-scale features are then fed into our fusion module and lightweight decoder for scene text prediction.

\subsection{Step-wise LoRA Module}

Scene text detection exhibits scale-dependent requirements: high resolution stage is critical for stroke boundaries and thin structures, middle stages govern instance connectivity and deformation consistency, and low-resolution stages mainly contribute global layout priors and background suppression. Following our SAM2-Hiera hierarchy defined in Section 3.2, we denote the Hiera blocks at stage \(k\) as shown in \cref{eq:k_scale}. We insert a step-wise LoRA module (SWLoRA) before each frozen Hiera block to perform a multi-step, low-rank residual refinement of the incoming token features. SWLoRA further employs a cosine-similarity criterion to dynamically stop the refinement once the representation stabilizes, allocating extra computation only to hard samples and hard stages.

For any block \(b_{k,i}\in\mathcal{B}_k\), let its frozen mapping be \(f_{k,i}(\cdot)\). With SWLoRA, the block output is computed as
\begin{equation}
\mathbf{y}_{k,i} = f_{k,i}\big(\mathrm{SWLoRA}_{k,i}(\mathbf{x}_{k,i})\big),
\label{eq:swlora_wrap}
\end{equation}
where \(\mathbf{x}_{k,i}\) and \(\mathbf{y}_{k,i}\) are the input and output token representations of the block  \(b_{k,i}\), respectively.

\subsubsection{Step-wise low-rank residual refinement.}
Instead of applying a single static LoRA update, SWLoRA performs up to \(T\) refinement steps. Let \(\mathbf{x}_{k,i}^{(0)}=\mathbf{x}_{k,i}\). The refinement process can be represented as
\begin{equation}
\mathbf{x}_{k,i}^{(t+1)} = \mathbf{x}_{k,i}^{(t)} + \gamma_{k,i,t}\,\Delta_{k,i}\!\left(\mathbf{x}_{k,i}^{(t)}\right),
\qquad t=0,\dots,T-1,
\label{eq:swlora_iter}
\end{equation}
where \(\gamma_{k,i,t}\) is a learnable step scale, and \(\Delta_{k,i}(\cdot)\) is a low-rank residual mapping:
\begin{equation}
\Delta_{k,i}(\mathbf{x}) =
\frac{\alpha}{r}\,
\mathbf{W}^{(k,i)}_{\mathrm{up}}
\Big(
\mathbf{W}^{(k,i)}_{\mathrm{down}}\big(\mathrm{Dropout}(\mathrm{LN}(\mathbf{x}))\big)
\Big).
\label{eq:swlora_delta}
\end{equation}
Here \(\mathrm{LN}(\cdot)\) is layer normalization, \(r\) is the LoRA rank, \(\alpha\) is the LoRA scaling factor, and \(\mathbf{W}^{(k,i)}_{\mathrm{down}},\mathbf{W}^{(k,i)}_{\mathrm{up}}\) are the down-/up-projection matrices. After refinement, SWLoRA outputs \(\mathbf{x}_{k,i}^{(T^\ast)}\) to the frozen block, where \(T^\ast\le T\) is the number of the executed steps.

\subsubsection{Cosine-similarity based dynamic early-exit.}
Cosine-similarity based dynamic early-exit mechanism adaptively determines the effective refinement depth \(T^\ast\) for each sample and each block. The key motivation is that the required refinement depth is not constant: it varies with input difficulty and with the stage/block characteristics. In experiments, we set the maximum number of refinement steps to \(T=5\).

Specifically, after obtaining \(\mathbf{x}_{k,i}^{(t+1)}\), we compute the cosine similarity between consecutive states:
\begin{equation}
\mathrm{cos}_{k,i}^{(t)} =
\frac{
\left\langle \mathrm{vec}\!\left(\mathbf{x}_{k,i}^{(t+1)}\right),\ \mathrm{vec}\!\left(\mathbf{x}_{k,i}^{(t)}\right)\right\rangle
}{
\left\|\mathrm{vec}\!\left(\mathbf{x}_{k,i}^{(t+1)}\right)\right\|_2
\left\|\mathrm{vec}\!\left(\mathbf{x}_{k,i}^{(t)}\right)\right\|_2
+\varepsilon
},
\label{eq:swlora_cos}
\end{equation}
where \(\mathrm{vec}(\cdot)\) flattens the token and channel dimensions and \(\varepsilon\) is a small constant for numerical stability. A high \(\mathrm{cos}_{k,i}^{(t)}\) indicates that the refinement update becomes directionally stable, suggesting that the current representation has already absorbed the necessary task-specific correction at this block.

Given a minimum-step constraint \(T_{\min}\)=1, we stop refinement early if
\begin{equation}
t+1\ge T_{\min} \quad \text{and} \quad \mathrm{cos}_{k,i}^{(t)}>\tau,
\label{eq:swlora_exit}
\end{equation}
where \(\tau\) is the exit threshold. Once \cref{eq:swlora_exit} is satisfied, we set \(T^\ast=t+1\) and output \(\mathbf{x}_{k,i}^{(T^\ast)}\) to the frozen block, otherwise, the refinement continues until reaching the maximum step \(T=5\).

This dynamic mechanism encourages SWLoRA to allocate refinement depth where it is truly needed, especially for hard samples or blocks requiring stronger correction, and to stop once the representation stabilizes, leading to more effective feature adaptation across samples and blocks.

\subsection{Common Subspace Fusion (CSF)}

Given the scale-aligned feature pairs $\{(S_k,D_k)\}_{k=1}^{4}$, we fuse them to combine structural cues from SAM2 encoding branch with domain-robust semantics from the DINOv3 branch.

\subsubsection{Channel alignment and gated mixing.}
We first project both branches to a common channel dimension $C$:
\begin{equation}
\hat S_k = \phi_s^{(k)}(S_k),\qquad \hat D_k = \phi_d^{(k)}(D_k),\qquad
\hat S_k,\hat D_k\in\mathbb{R}^{B\times C \times H_k \times W_k},
\label{eq:csf_align}
\end{equation}
where $\phi_s^{(k)}$ and $\phi_d^{(k)}$ are lightweight $1\times 1$ convolutions. We then compute an input-adaptive gate $\alpha_k$ from global statistics and form an intermediate mixture:
\begin{equation}
\alpha_k=\sigma\!\Big(\mathrm{MLP}\big([\mathrm{GAP}(\hat S_k);\mathrm{GAP}(\hat D_k)]\big)\Big),\qquad
M_k=\alpha_k\odot \hat S_k + (1-\alpha_k)\odot \hat D_k,
\label{eq:csf_gate}
\end{equation}
where $\sigma$ represents the Sigmoid function, $\alpha_k$ is a scalar or channel-wise weight, $\odot$ denotes element-wise multiplication, and
$\mathrm{GAP}(\cdot)$ is global average pooling.

\subsubsection{Common subspace estimation.}
Let $N_k=H_kW_k$ and denote by $X_{S_k},X_{D_k}\in\mathbb{R}^{B\times C\times N_k}$ the spatially flattened,
mean-centered features of $\hat S_k$ and $\hat D_k$, respectively. We estimate per-sample covariances and
construct a second-order proxy for shared variation:
\begin{equation}
\Sigma_{S_k}=\frac{1}{N_k}X_{S_k}X_{S_k}^{\top}+\varepsilon \mathbf{I},\quad
\Sigma_{D_k}=\frac{1}{N_k}X_{D_k}X_{D_k}^{\top}+\varepsilon \mathbf{I},\quad
\end{equation}
\begin{equation}
A_k=\Sigma_{S_k}\Sigma_{D_k},
\end{equation}
where $\varepsilon \mathbf{I}$ is added for numerical stability. We approximate the top-$r$ eigenspace (r=16) of $A_k$ and obtain an orthonormal basis
$Q_k\in\mathbb{R}^{B\times C\times r}$.

\subsubsection{Fusion in the common subspace with DINO complement preservation.}
Given $Q_k$, we define the projection onto the estimated common subspace:
\begin{equation}
\mathcal{P}_k(X)=Q_k(Q_k^{\top}X),
\label{eq:csf_proj}
\end{equation}
applied to flattened features and reshaped back to $\mathbb{R}^{B\times C\times H_k\times W_k}$.
We restrict cross-branch interaction to the projected component, while explicitly preserving DINOv3's
orthogonal complement:
\begin{equation}
F_k=\mathcal{P}_k(M_k)\;+\;\big(\hat D_k-\mathcal{P}_k(\hat D_k)\big),
\qquad F_k\in\mathbb{R}^{B\times C\times H_k\times W_k}.
\label{eq:csf_fuse}
\end{equation}
The first term enforces fusion only on the statistically shared subspace, whereas the second term prevents
the fusion operator from implicitly removing DINOv3-specific components, and feed $\{F_k\}_{k=1}^{4}$ to the decoder for multi-scale prediction.

\section{Experiments}

\subsection{Datasets}

\subsubsection{Normal-condition datasets.} 
\textbf{CTW-1500 \cite{yuliang2017detecting}} is a benchmark focusing on curved text with polygon-based annotations for long and irregular text lines. Following the standard split, it contains 1000 training images and 500 testing images. \textbf{Total-Text \cite{ch2017total}} covers multi-oriented and curved text in natural scenes and is widely used to evaluate localization under diverse layouts. It provides 1255 training images and 300 testing images in the commonly adopted split. \textbf{MLT \cite{nayef2017icdar2017}} (Multi-Lingual Text) targets multi-lingual, multi-script scene text with substantial appearance variation. Here we utilize 7200 training set images and 1800 images for testing. 

\subsubsection{Adverse-condition datasets.} To systematically evaluate robustness under real-world imaging degradations, we construct degraded variants of CTW-1500, Total-Text, and MLT. Specifically, we create five subsets of \textbf{Rain}, \textbf{Fog}, \textbf{Underexposure}, \textbf{Overexposure}, and \textbf{Low-resolution} by applying corresponding degradation processes to original images while keeping ground-truth annotations unchanged. Detailed examples are provided in the supplementary material. We also conducted zero-shot experiments on low-light scenes under real nighttime conditions, with details in the supplementary materials.

\subsection{Implementation Details}

We implement our proposed TextDS in PyTorch and initialize it with two pretrained backbones: SAM2-Hiera-L and DINOv3 ViT-L/16. During training, images and corresponding masks are resized to $512\times512$. All experiments are conducted on a single NVIDIA RTX 4090 GPU. We train for 50 epochs with a batch size of 4 using the AdamW optimizer, with an initial learning rate of 0.001 and weight decay of $5\times10^{-4}$. The learning rate is scheduled with cosine annealing and a minimum learning rate of $1\times10^{-7}$. We optimize a structure loss function composed of an equal-weighted binary cross-entropy term and an IoU term to emphasize boundary regions.

\subsection{Evaluation Metrics}

We evaluate detection performance using Precision (P), Recall (R), and F-measure (F). Precision measures the fraction of predicted positives that are correct, while recall measures the fraction of ground-truth positives that are successfully detected. Considering both precision and recall comprehensively, we further use F-measure as the comprehensive indicator.

\begin{table}[t]
\centering
\caption{Comparison of scene text detection accuracy and model complexity. Accuracy metrics include Precision (P), Recall (R), and F-measure (F) (\%). Model complexity includes whether SynthText pre-training (Synth) is required, the number of trainable parameters (Param.), and FPS (uniformly measured on the same RTX 4090 GPU).}
\setlength{\tabcolsep}{2.2pt}
\renewcommand{\arraystretch}{1.0}

\begin{adjustbox}{max width=\textwidth}
\begin{tabular}{l|c|c|c|ccc|ccc|ccc|c}
\toprule
\multirow{2}{*}{Method} & \multirow{2}{*}{Published} & \multirow{2}{*}{Synth} & \multirow{2}{*}{Param. (M)}
& \multicolumn{3}{c|}{CTW-1500} & \multicolumn{3}{c|}{Total-Text} & \multicolumn{3}{c|}{MLT}
& \multirow{2}{*}{FPS} \\
\cline{5-13}
 &  &  &  & P & R & F & P & R & F & P & R & F &  \\
\midrule
DB-Net   \cite{liao2020real}       & AAAI 2020  & \checkmark & 28.0 & 86.9 & 80.2 & 83.4 & 87.1 & 82.5 & 84.7 & 87.8 & 83.5 & 85.6 & 35.0 \\
TextBPN  \cite{zhang2021adaptive}  & ICCV 2021  & \checkmark & 38.7 & 86.5 & 83.6 & 85.0 & 90.7 & 85.2 & 87.9 & 90.3 & 84.6 & 87.4 & 22.6 \\
TextPMs  \cite{zhang2022arbitrary} & TPAMI 2022 & \checkmark & 36.4 & 87.8 & 83.8 & 85.7 & 90.6 & 86.5 & 88.5 & 88.1 & 83.6 & 85.8 & 16.8 \\
TextDCT  \cite{su2022textdct}      & TMM 2023   & \checkmark & -    & 85.0 & 85.3 & 85.1 & 87.2 & 82.7 & 84.9 & - & - & - & - \\
CT-Net   \cite{shao2023ct}         & TCSVT 2023 & \checkmark & -    & 87.9 & 82.7 & 85.2 & 90.8 & 85.0 & 87.8 & - & - & - & - \\
LRANet   \cite{su2024lranet}       & AAAI 2024  & \checkmark & 34.3    & 89.4 & 85.5 & 87.4 & 88.9 & 86.6 & 87.7 & 92.7 & 84.4 & 88.3 & 38.1 \\
CB-Net   \cite{zhao2024cbnet}      & IJCV 2024  & \checkmark & 12.3 & 89.3 & 82.9 & 86.0 & 90.1 & 82.5 & 86.1 & \textbf{92.9} & 85.2 & 88.9 & 32.9 \\
ERRNet   \cite{su2025explicit}     & AAAI 2025  & \checkmark & -    & 88.1 & 85.3 & 86.7 & 90.1 & 86.1 & 88.1 & - & - & - & - \\
S3INet   \cite{wang2025s3inet}     & TNNLS 2025 & \checkmark & 38.0 & 89.2 & 83.0 & 86.0 & \textbf{91.2} & 86.2 & 88.7 & 92.8 & 86.8 & 89.7 & 37.3 \\
\midrule
TextDS                           & -          & \ding{55}   & \textbf{4.9} 
& \textbf{91.6} & \textbf{89.6} & \textbf{90.6} & 89.7 & \textbf{88.4} & \textbf{89.1} & 92.4 & \textbf{92.0} & \textbf{92.2} & \textbf{44.1} \\
\bottomrule
\end{tabular}
\end{adjustbox}

\label{tab:PRF_complexity}
\end{table}

\begin{figure}[!htbp]
  \centering
  \includegraphics[height=4.3cm]{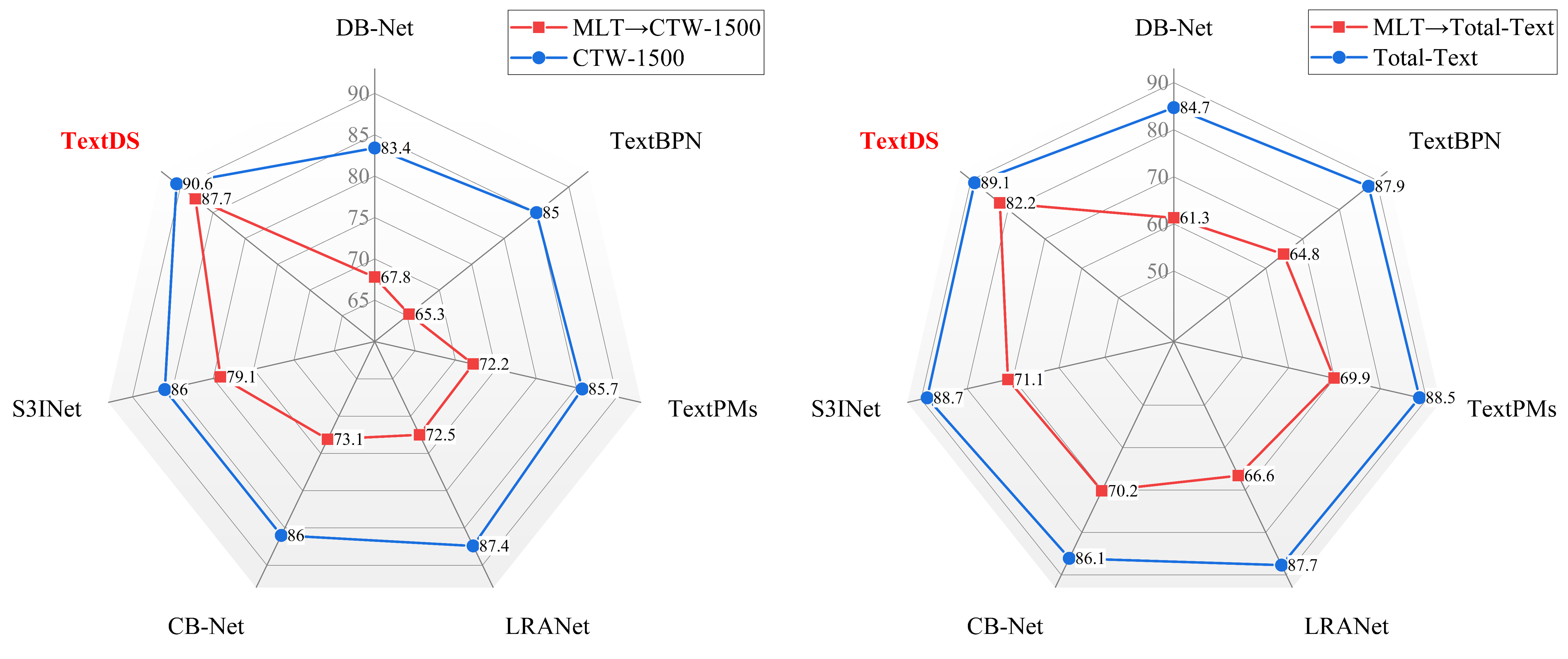}
  \caption{The results of domain generalization from the MLT dataset to CTW-1500 and Total-Text are compared between TextDS and the comparison methods. The blue circles represent the F-measure of the model on CTW-1500 and Total-Text by itself, while the red circles represent the F-measure of the model when generalizing from MLT dataset to CTW-1500 and Total-Text.
  }
  \label{fig3}
\end{figure}

\subsection{Comparison with State-of-the-art Methods}

Our experimental results on the CTW-1500, Total-Text and MLT datasets under normal conditions without distribution shifts are shown in \cref{tab:PRF_complexity}. The methods used for comparison include those from the representative approaches in recent years, including DB-Net \cite{liao2020real}, TextBPN \cite{zhang2021adaptive}, TextPMs \cite{zhang2022arbitrary}, TextDCT \cite{su2022textdct}, CT-Net \cite{shao2023ct}, LRANet \cite{su2024lranet}, CB-Net \cite{zhao2024cbnet}, ERRNet \cite{su2025explicit}, and S3INet \cite{wang2025s3inet}. The TextDS we proposed significantly outperforms the comparison methods in terms of detection performance in Precision, Recall, and F-measure. Especially, it is notably superior to the methods developed in recent two years (2024 and 2025), indicating the outstanding performance of TextDS on the in-distribution cases.

\subsection{Comparison of Model Complexity}

While maintaining accuracy beyond existing models, the TextDS we propose, due to its Data- and Parameter-efficient structural design, requires a detailed analysis and comparison of its computational complexity. Model complexity comparison includes whether SynthText \cite{gupta2016synthetic} pre-training is required, the number of trainable parameters, and the Frames Per Second (FPS) value. The specific results are shown in \cref{tab:PRF_complexity}. To ensure rigor, results that were not code-open-sourced and not reported in the past paper are indicated by (-). Compared with the comparison methods, the characteristics of TextDS make it no need to pre-train with the specific large-scale SynthText dataset for scene text detection. On this basis, TextDS only requires 4.9M parameters, significantly lower than the comparison methods. Moreover, while maintaining the same hardware configuration, TextDS achieves the highest 44.1 FPS value, demonstrating the efficiency of our design and providing a new approach suitable for practical deployment.

\subsection{Experimental Results of Domain Generalization}

The domain generalization results of TextDS and the comparison methods from the MLT dataset to CTW-1500 and Total-Text are shown in \cref{fig3}. The comparison methods include DB-Net \cite{liao2020real}, TextBPN \cite{zhang2021adaptive}, TextPMs \cite{zhang2022arbitrary}, LRANet \cite{su2024lranet}, CB-Net \cite{zhao2024cbnet}, and S3INet \cite{wang2025s3inet}. The blue circles represent the in-domain F-measure of the models on CTW-1500 and Total-Text, while the red circles represent the F-measure of the models when generalizing from MLT dataset to CTW-1500 and Total-Text. It can be intuitively observed that TextDS, when performing the best in domain-specific data performance, also exhibits significantly superior performance in domain generalization of different scene text detection datasets.

\begin{table}[tb]
\centering
\caption{Experimental results of our proposed degraded scene text image datasets under adverse conditions. Imaging conditions include weather (rain and fog), exposure (underexposure and overexposure), and low-resolution (256×256 and 128×128).}
\setlength{\tabcolsep}{3pt}
\renewcommand{\arraystretch}{0.85}
\setlength{\aboverulesep}{0.2ex}
\setlength{\belowrulesep}{0.2ex}

\begin{tabular}{l|l|ccc|ccc|ccc}
\toprule
\multicolumn{2}{l|}{\multirow{2}{*}{Imaging Condition}}  &
\multicolumn{3}{c|}{CTW-1500} &
\multicolumn{3}{c|}{Total-Text} &
\multicolumn{3}{c}{MLT} \\
\cline{3-11}
\multicolumn{2}{l|}{} &
\rule{0pt}{2ex}P & R & F &
\rule{0pt}{2ex}P & R & F &
\rule{0pt}{2ex}P & R & F \\
\midrule
\multicolumn{2}{l|}{Normal} &
91.6 & 89.6 & 90.6 &
89.7 & 88.4 & 89.1 &
92.4 & 92.0 & 92.2 \\
\midrule
\multirow{2}{*}{Weather} & Rain &
90.2 & 88.3 & 89.2 &
87.3 & 88.2 & 87.7 &
92.4 & 90.9 & 91.6 \\
 & Fog &
91.7 & 88.0 & 89.8 &
88.4 & 88.2 & 88.3 &
92.5 & 91.4 & 91.9 \\
\midrule
\multirow{2}{*}{Exposure} & Underexposure &
91.3 & 88.9 & 90.1 &
87.6 & 88.0 & 87.8 &
92.0 & 92.2 & 92.1 \\
 & Overexposure &
89.9 & 90.0 & 89.9 &
88.0 & 87.9 & 87.9 &
91.9 & 91.7 & 91.8 \\
\midrule
\multirow{2}{*}{Low-resolution} & 256$\times$256 &
90.2 & 89.3 & 89.8 &
88.7 & 87.9 & 88.3 &
90.8 & 91.1 & 90.9 \\
 & 128$\times$128 &
88.4 & 89.4 & 88.9 &
87.0 & 89.2 & 88.0 &
86.9 & 89.9 & 88.3 \\
\bottomrule
\end{tabular}
\label{tab:degraded}
\end{table}

\begin{figure}[!htbp]
  \centering
  \includegraphics[height=5.3cm]{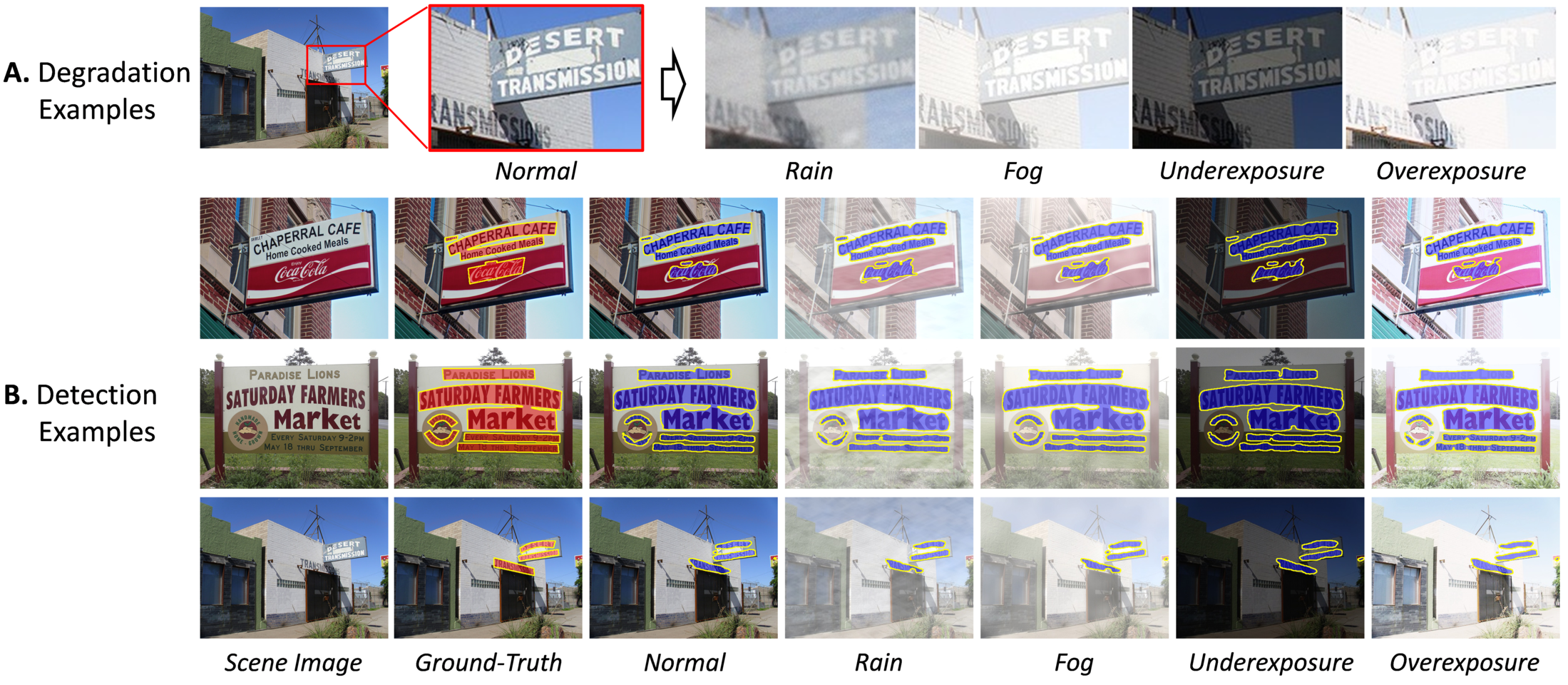}
  \caption{Visualization of the detailed scene text image degradation and detection examples for TextDS. For the detection samples, the red color represents the Ground-Truth, while the blue color represents the text region results detected by the TextDS, which maintains performance under adverse imaging conditions.
  }
  \label{fig4}
\end{figure}

\begin{figure}[tb]
  \centering
  \includegraphics[height=3.45cm]{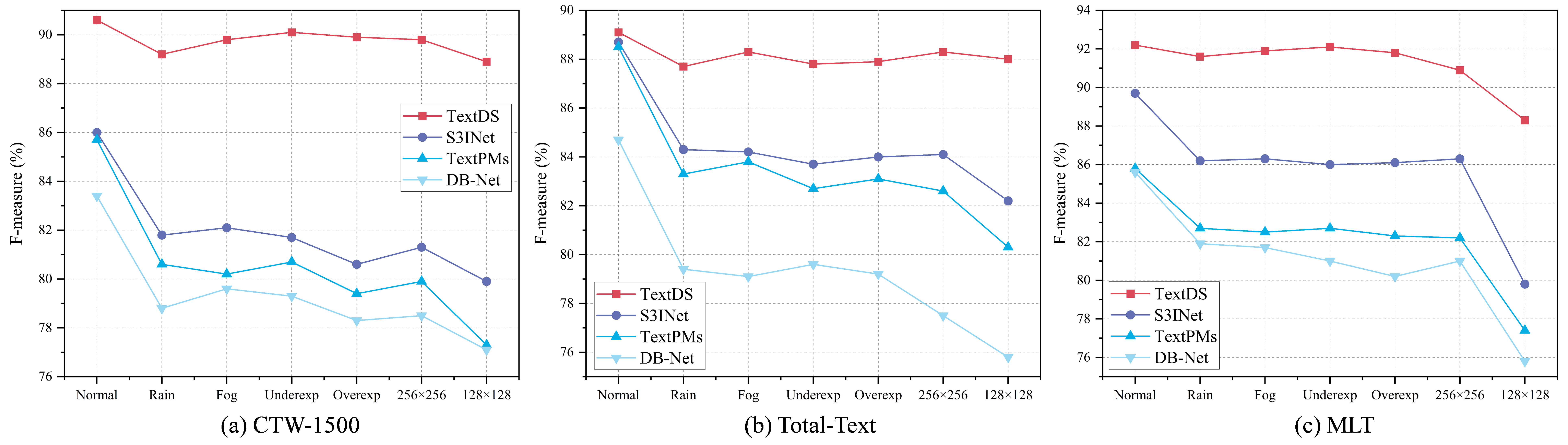}
  \caption{Performance comparison of TextDS and the comparison methods on the scene text datasets in adverse-condition scenarios, including the F-measure of TextDS, S3INet, TextPMs and DBNet under Normal and degraded imaging conditions.
  }
  \label{fig5}
\end{figure}

\begin{table}[tb]
\centering
\caption{Ablation experimental results of the key components on F-measure (\%).}
\setlength{\tabcolsep}{4pt}
\renewcommand{\arraystretch}{0.6}
\begin{tabular}{ccc|ccc}
\toprule
Dual-Encoding & SWLoRA & CSF & CTW-1500 & Total-Text & MLT \\
\midrule
  &   &   & 82.2 & 85.1 & 86.4 \\
\checkmark &   &   & 85.3 & 86.0 & 88.1 \\
\checkmark & \checkmark &   & 87.9 & 88.3 & 90.3 \\
\checkmark &   & \checkmark & 88.2 & 87.8 & 91.0 \\
\checkmark & \checkmark & \checkmark & \textbf{90.6} & \textbf{89.1} & \textbf{92.2} \\
\bottomrule
\end{tabular}
\label{tab:ablation}
\end{table}


\begin{figure}[!htbp]
  \centering
  \includegraphics[height=3.6cm]{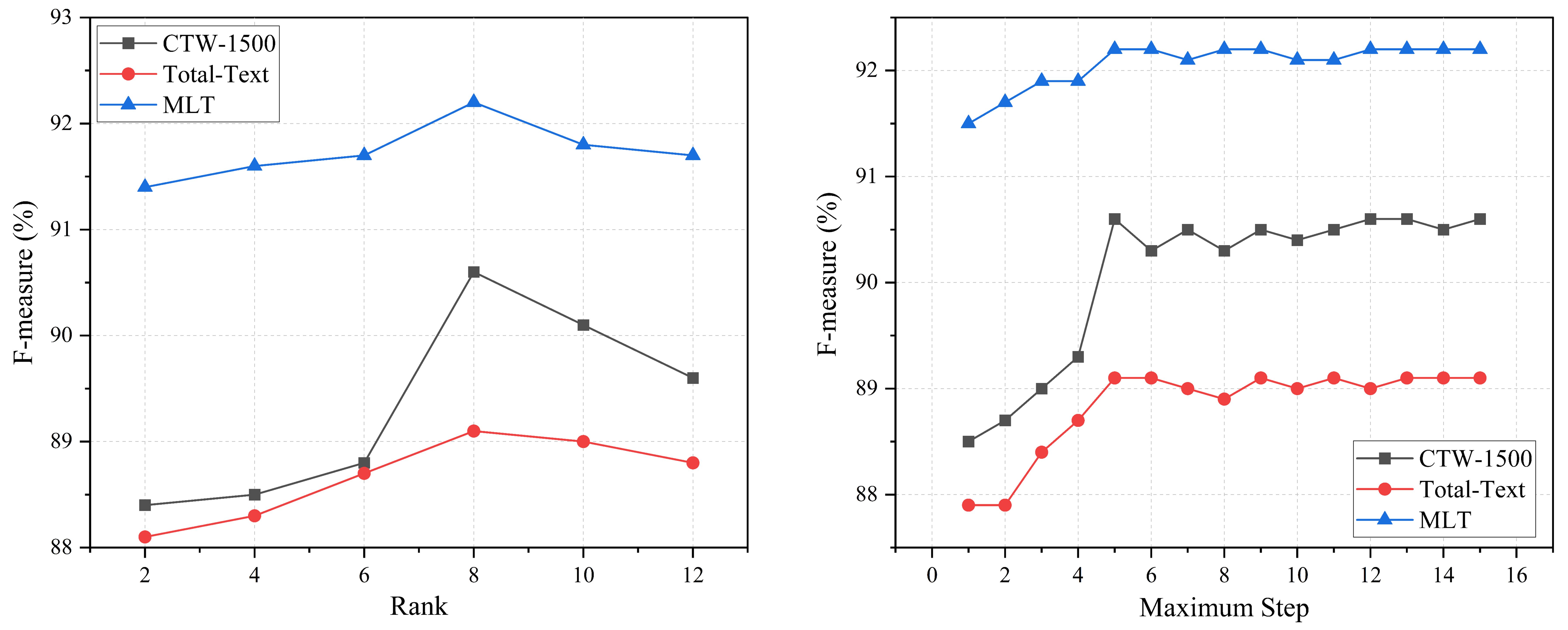}
  \caption{Variation of the F-measure with the selection of Rank and Maximum Step, using the SWLoRA structure to fine-tune CTW-1500, Total-Text, and MLT datasets, where Rank $=8$ and Maximum Step $=5$ are adopted as the default setting.
  }
  \label{fig6}
\end{figure}

\subsection{Experimental Results of Adverse-condition Datasets}

The experimental results of the degraded dataset under adverse conditions are shown in \cref{tab:degraded}, and the visualization is of degradation examples and detection results is shown in  \cref{fig4}. Based on the CTW-1500, Total-Text and MLT datasets, we processed and obtained their respective datasets of adverse imaging conditions, including weather (rain and fog), exposure (underexposure and overexposure) and low-resolution (256 × 256 and 128 × 128). It can be seen that TextDS has demonstrated robustness when dealing with these degraded datasets. This part of the experiment fills the gap in the datasets for scene text detection under adverse imaging conditions. \cref{fig5} compares the performance of TextDS with several representative baseline methods on the adverse-condition scene text dataset, reporting the F-measure of TextDS, S3INet \cite{wang2025s3inet}, TextPMs \cite{zhang2022arbitrary}, and DBNet \cite{liao2020real} under the normal setting and various degraded imaging conditions. As can be clearly observed, under degraded conditions, TextDS achieves consistently higher overall performance than the competing methods and exhibits a clearly smaller drop relative to the F-measure of normal setting, demonstrating its strong robustness to adverse imaging conditions.


\subsection{Ablation Studies and Discussion}

\subsubsection{Ablation experiments of Key components.} The ablation experimental results of TextDS are shown in \cref{tab:ablation}, using F-measure as the evaluation metric. The key components include the Dual-Encoding, SWLoRA and CSF modules. The ablation experiments further demonstrated the effectiveness of the proposed separate and combined use of the structures. Detailed ablation experiments for datasets under adverse conditions are presented in the supplementary materials.


\subsubsection{Selection of Rank and Step.}
Using the SWLoRA structure for fine-tuning, we further study how the final performance depends on two key hyper-parameters: the LoRA rank and the maximum step in SWLoRA. As shown in \cref{fig6}, increasing the Rank or Maximum Step generally improves the fitting ability at the beginning, but the gains quickly saturate and may become unstable when the adaptation capacity is overly large. Considering the overall trend across the three datasets, as well as parameter efficiency and convergence stability, we set Rank $=8$ and Maximum Step $=5$ as the default configuration for all experiments. Under this setting, the proposed early-exit mechanism further reduces unnecessary refinement computation, with average executed steps of 3.652, 2.205, and 3.476 out of 5 on CTW-1500, Total-Text, and MLT, respectively, corresponding to reductions of 27.0\%, 55.9\%, and 30.5\%. These results show that SWLoRA can adaptively stop the refinement process once the representation becomes sufficiently stable, achieving a favorable balance between accuracy, stability, and computational efficiency.

\subsubsection{Comparison with LLM-based and end-to-end OCR methods.} 
\begin{figure}[tb]
  \centering
  \includegraphics[height=1.6cm]{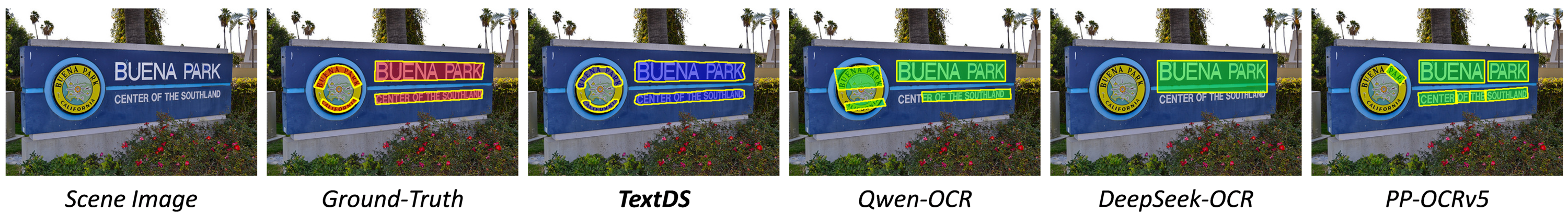}
  \caption{Visual comparison with LLM-based and end-to-end OCR methods in scene text detection, including Qwen-OCR, DeepSeek-OCR, and PP-OCR.
  }
  \label{fig7}
\end{figure}

We further conducted a visual qualitative comparison between proposed TextDS and representative LLM-based and end-to-end OCR methods, including Qwen-OCR \cite{bai2025qwen3}, DeepSeek-OCR \cite{wei2025deepseek}, and PP-OCR \cite{li2022pp}, as shown in \cref{fig7}. The LLM-based methods have strong capabilities in semantic understanding \cite{sun2025edvd}, but they have weaknesses in pixel-level geometric positioning of complex deformed texts. TextDS, as a front-end detector, due to its extremely strong structural prior, is highly complementary to the existing LLM-VLM models and can provide higher-quality text extraction regions for large language models. More comparison details and examples can be found in the supplementary material.





\section{Conclusion}

This paper addresses the practical challenge that scene text detectors face distribution shifts and real-world imaging degradations at deployment. We propose TextDS, an efficient scene text detection network that uses a data-efficient dual-encoder strategy combining SAM2 and DINOv3 to capture complementary text-related representations without large-scale scene-text-specific pretraining, and introduces Step-wise LoRA Adaptation and Common Subspace Fusion for parameter-efficient fine-tuning and alignment of the two encoding branches. In addition, we build adverse-condition scene text detection subsets to enable systematic evaluation under imaging degradations. Extensive experiments show that TextDS achieves competitive detection performance while improving robustness in domain generalization and adverse imaging conditions.

\section*{Acknowledgements}
The research work described in this paper was conducted in the JC STEM Lab of Machine Learning and Computer Vision funded by The Hong Kong Jockey Club Charities Trust. This research received partially support from the Global STEM Professorship Scheme from the Hong Kong Special Administrative Region.

%
%
\bibliographystyle{splncs04}
\bibliography{main}
\end{document}